\documentclass[5p,twocolumn]{elsarticle}
\usepackage{epsfig}
\usepackage{multirow}
\usepackage{setspace}
\usepackage{amssymb,amsmath,amsfonts}
\usepackage{hyperref}
\usepackage{makecell}
\usepackage{todonotes}
\usepackage{placeins}
\usepackage{babel,blindtext}

\usepackage{amssymb}
\begin{document}

\begin{frontmatter}

\title{Mobile Behavioral Biometrics for Passive Authentication}

\author[inst1]{Giuseppe Stragapede\corref{cor1}}
\ead{giuseppe.stragapede@uam.es}

\affiliation[inst1]{organization={Biometrics and Data Pattern Analytics Lab, Universidad Autonoma de Madrid},%Department and Organization
            % addressline={}, 
            % city={City One},
            % postcode={00000}, 
            % state={State One},
            country={Spain}}
\cortext[cor1]{Corresponding author.}

\affiliation[inst2]{organization={Orange Labs},%Department and Organization
            % addressline={}, 
            city={Cesson-Sevigne},
            % postcode={00000}, 
            % state={State One},
            country={France}}

\author[inst1]{Ruben Vera-Rodriguez}
\ead{ruben.vera@uam.es}
\author[inst1]{Ruben Tolosana}
\ead{ruben.tolosana@uam.es}
\author[inst1]{\\Aythami Morales}
\ead{aythami.morales@uam.es}
\author[inst1]{Alejandro Acien}
\ead{alejandro.acien@uam.es}
\author[inst2]{Ga\"el Le Lan}
\ead{gael.lelan@orange.com}
% \affiliation[inst2]{organization={Department Two},%Department and Organization
%             addressline={Address Two}, 
%             city={City Two},
%             postcode={22222}, 
%             state={State Two},
%             country={Country Two}}

\begin{abstract}
%% Text of abstract
Current mobile user authentication systems based on PIN codes, fingerprint, and face recognition have several shortcomings. Such limitations have been addressed in the literature by exploring the feasibility of passive authentication on mobile devices through behavioral biometrics. 
% \par The\fi the 
In this line of research, this work carries out a comparative analysis of unimodal and multimodal behavioral biometric traits acquired while the subjects perform different activities on the phone such as typing, scrolling, drawing a number, and tapping on the screen, considering the touchscreen and the simultaneous background sensor data (accelerometer, gravity sensor,
gyroscope, linear accelerometer, and magnetometer). 
Our experiments are performed over HuMIdb\footnotemark, one of the largest and most comprehensive freely available mobile user interaction databases to date. A separate Recurrent Neural Network (RNN) with triplet loss is implemented for each single modality. Then, the weighted fusion of the different modalities is carried out at score level. 
In our experiments, the most discriminative background sensor is the magnetometer, whereas among touch tasks the best results are achieved with keystroke in a fixed-text scenario. In all cases, the fusion of modalities is very beneficial, leading to Equal Error Rates (EER) ranging from 4\% to 9\% depending on the modality combination in a 3-second interval.

\end{abstract}

% %%Research highlights
% \begin{highlights}
% \item We present a first benchmark of HuMIdb, a publicly available HCI database.
% \item We evaluate touchscreen and background sensor data for mobile passive authentication.
% \item We train a recurrent neural network with triplet loss for each individual modality.
% \item A multimodal system is achieved through the fusion of modalities at score level.
% \item Our results show the suitability of behavioral biometrics for mobile authentication.
% \end{highlights}

\end{frontmatter}

\section{Introduction}
\label{sec:intro}
\footnotetext{\url{https://github.com/BiDAlab/HuMIdb}}

\par Most current mobile authentication systems are based on the knowledge of \textit{secrets}, as in the case of passwords, PIN codes or graphic patterns, or on physiological biometrics\footnote{The biological characteristics suited for identifying an individual are referred to as physiological biometrics.}, as for fingerprint or face recognition systems. 
\par Such systems, however, present several limitations in terms of security and usability \cite{JAIN201680}. The former are in fact exposed to \textit{shoulder-surfing} attacks, which consist in the attacker peeking at the screen of the unaware authenticating user, or \textit{guess} attacks (users tend to choose simple secret sequences, i.e. their birth dates, exploiting only a small set of all possible PIN codes or passwords). Additionally, with special lighting and high-resolution cameras, oily residues on screen surfaces can reveal secret patterns (\textit{smudge} attacks) \cite{Aviv2010}.
Furthermore, studies have demonstrated how on average mobile users spend about 2.9\% of their total usage time for knowledge-based authentication \cite{Zezschwitz2016}. The latter, in turn, are inclined to \textit{presentation} attacks (spoofing), albeit they do not involve mnemonic efforts \cite{2019_TIFS_Fingerprint_PAs_Tolosana, marcel2019handbook}. 
In any case, they are both not suited for offering prolonged protection throughout the entire device usage time. In such case, users would in fact have to repeatedly interrupt their activity to carry out the authentication process, for instance by placing their fingertip on the dedicated scanner or by typing in their PIN code. On the other hand, frequent face verification seems infeasible due to hardware constraints, such as the computational overload, memory overhead and battery consumption of the acquisition and processing of images. 
Consequently, if an attacker gains access to the device, they can stay authenticated as long as it remains active. 

\par On the other hand, Continuous Authentication\footnote{The terms \textit{continuous}, \textit{implicit}, and \textit{transparent} authentication appear interchangeably in the literature \cite{7503170}.} (CA) approaches are designed to constantly verify the biometric features of the user in a \textit{passive} way, in other words, without having them to carry out any specific authentication task \cite{7503170}. CA on mobile devices can be achieved by employing behavioral biometrics. All the means that enable or contribute to differentiating between individuals throughout the \textit{way} they perform activities such as gait, typing, scrolling, etc., can be ascribed to such category \cite{Acien2021}. {\color{black} They have been also explored in conjunction with physiological biometrics in particular circumstances, such as placing a call \cite{7927733}, or during the entry-point authentication process \cite{7139043}.}
\par It has been pointed out that behavioral biometrics offer stronger security guarantees compared to physiological biometrics, as spoofing requires more advanced technical skills \cite{2021_Arxiv_SVConGoing_Tolosana}. 
Additionally, at a cost of increased complexity, biometric systems typically benefit from the combination of different \textit{modalities}\footnote{In the context of user authentication, a \textit{modality} is a source of biometric information that allows to identify an individual.} in terms of robustness, immunity to noise, universality, and security \cite{7503170, 2018_INFFUS_MCSreview1_Fierrez, ROSS20032115}. Behavioral biometrics traits are suitable to achieve a mobile multimodal authentication system as mobile devices are equipped with several sensors, capable of simultaneously acquiring a vast amount of behavioral biometric information{\color{black} , which can also reveal a significant amount of information about the user \cite{delgadosantos2021survey, 9320271, 8545797}}.

{\color{black}
\subsection{Related Work}
\label{sec:related_work}}

\par To date, behavioral biometrics for CA on mobile devices has been an active field of research for over a decade. Throughout the years, several studies have focused on the modalities explored within our work, employing a variety of different classification algorithms (mobile keystroke: \cite{maiorana2021mobile, 8576226}; touch data: \cite{de2021air, 8353868, tolosana_biotouchpass2, 2020_CDS_HCIsmart_Acien}; background sensor data: \cite{EhatishamulHaq2018}; multimodal biometrics: \cite{2019_MULEA_MultiLock_Acien, 6980382}).
Regarding touch data information it is worth mentioning several studies in the literature. Taking into account keystroke gestures, Acien \textit{et al.} proposed a Long Short-Term Memory (LSTM) RNN network for authentication at large scale in free-text scenarios, evaluating different loss functions (softmax, contrastive, and triplet loss), number of gallery samples, length of the keystroke sequences, and device type (physical vs touchscreen keyboard) \cite{Acien2021}. They obtained an EER of 9.2\% for touchscreen keyboards. Such Deep Learning (DL) architecture has in fact proved to work well with data consisting in low-dimensional time domain signals \cite{MAIORANA2020374}. In \cite{2020_CDS_HCIsmart_Acien}, swiping gestures were examined over the HuMIdb database. A system based on Siamese RNN architecture and 29 features extracted achieved an EER of 19\%. Handwriting recognition in a mobile context was taken into account in \cite{tolosana_biotouchpass2}. By combining Dynamic Time Warping (DTW) with a RNN architecture and evaluating one single character, an EER score of 11.90\% was achieved. This was reduced to 2.38\% in the case of considering four-character sequences. By combining tapping with accelerometer, after extracting 127 features, EERs down to 3.65\% were achieved on a self-collected database involving over 80 subjects in \cite{6980382}. 
\par Deb \textit{et al.} proposed a contrastive loss-based Siamese LSTM RNN architecture for multimodal passive authentication, where subjects can be verified without any explicit authentication step \cite{Deb2019}. They evaluated 8 modalities, namely, keystroke, GPS location, accelerometer, gyroscope, magnetometer, linear accelerometer, gravity, and rotation sensors on a small self-collected dataset comprised of measurements from 30 smartphones for 37 subjects. By performing the fusion of the scores achieved by the different modalities, they achieved 96.47\% of True Acceptance Rate (TAR) at a False Acceptance Rate (FAR) of 0.1\% considering 3-second time intervals for authentication. 
\par More recently, Abuhamad \textit{et al.} developed a similarly DL-based continuous authentication system based on smartphone sensors. The dataset was self-collected by the means of a mobile application users had to install on their phone and it involved 84 subjects in total. Their approach was based on different RNN LSTM architectures and on data-level fusion involving three sensors (accelerometer, gyroscope, and magnetometer). Within 1-second time intervals, an EER of 0.41\% was obtained \cite{Abuhamad}.

\begin{figure*}[ht]
	\centering
    \includegraphics[width=\linewidth]{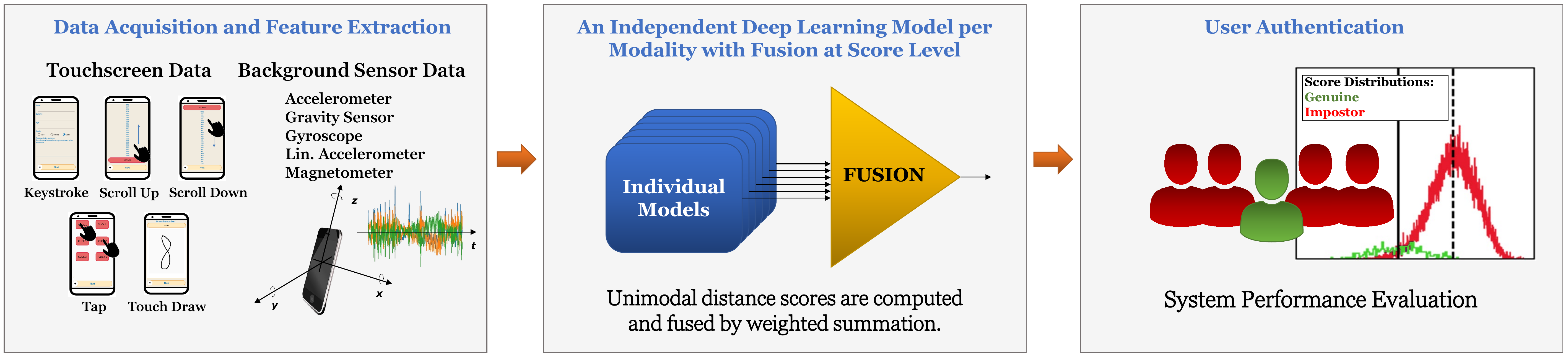}
    \vspace{-4mm}
    \caption{For each of the modalities, an independent network is trained and a multimodal system is achieved by fusion at score level (touchscreen data acquired during different tasks, and simultaneous background sensor data).}
    \label{fig:summary}
    \vspace{-2mm}

\end{figure*}

\vspace{.5cm}
{\color{black}
\subsection{System Overview}
\label{sec:overview}}

In this work we focus on mobile behavioral biometrics for passive authentication while the subjects interact with their own smartphone. For this we make use of HuMIdb database, in which touch and background sensor data is captured while subjects perform regular mobile interaction activities (keystroke, scroll up, scroll down, drawing a number with the finger, tap on the screen). Aiming to advance some aspects of the most recent and promising related studies described above (Sec. \ref{sec:related_work}), the main contributions of the current work are:
\begin{itemize}
    \item An in-depth analysis of individual behavioral biometric traits suitable for the application of CA through the interaction with mobile devices. In comparison with previous studies \cite{2019_MULEA_MultiLock_Acien, Deb2019, Abuhamad}, this work analyzes and proposes learned models based on: \textit{i)} specific tasks based on the most common user-device interaction activities. In this context, in fact, in relation to the systems' ability to discriminate among user data, it is crucial to differentiate between the notion of \textit{user} and \textit{device} \cite{Neverova2016, das2015exploring}. Our goal in this sense is to maximize the amount of biometric information in the raw data by analyzing data acquired while the user is carrying out dedicated acquisition sessions consisting in simple tasks designed to mimic the most common traits of mobile interaction, in a totally uncontrolled way. In contrast, on the one hand, gesture-based related studies take into account unimodal systems \cite{Acien2021, maiorana2021mobile, 8576226, de2021air, 8353868, tolosana_biotouchpass2, 2020_CDS_HCIsmart_Acien}; on the other hand, related studies based on DL models for multimodal behavioral biometrics do not have a specific focus on human gestures \cite{Deb2019, Abuhamad}; \textit{ii)} we analyze the information captured by the touchscreen in combination with simultaneous background sensor data to exploit the complementarity between task-dependent features and background sensors features (accelerometer, gravity sensor, gyroscope, linear accelerometer, magnetometer).
    \item First benchmark of HuMIdb (Human Mobile Interaction database), a novel and public database comprising more than 5GB from a wide range of mobile sensor data acquired under unsupervised scenario for user passive authentication \cite{Acien2020b}. Differently from related studies, the selected database is public and collected from a higher number of data subjects. Additionally, we make our benchmark publicly available\footnote{\url{https://github.com/BiDAlab/HuMIdbBenchmark}} to the scientific community for reproducibility and to promote additional research. 
    \item Implementation, for each individual modality, of a separate LSTM RNN. For training, we employ the triplet loss function (Sec. \ref{subsec:triplet_loss}) rather than the contrastive loss function (based on pairs, and used in \cite{Deb2019}) or binary cross-entropy (used in \cite{Abuhamad}) as our preliminary experiments on the HuMIdb showed it to be more effective. We compare the biometric performance of individual modalities and their fusion at score level. Moreover, with regard to the background sensors, we adopt a much higher sampling frequency than in \cite{Deb2019} (50Hz instead of 1Hz), and we make use of more features, such as first- and second-order derivatives (Sec. \ref{subsec:background}).
    \item Fusion of biometric and background sensor information, achieving accurate results ranging from 4\% to 9\% EER. In comparison with \cite{Abuhamad}, we adopt score level fusion rather than data level fusion. Such fusion strategy is discussed in Sec. \ref{subsec:fusion_of_modalities}.
\end{itemize}

Fig. \ref{fig:summary} represents the general architecture of the system proposed in this work.

\section{System Description}
\label{sec:system_description}

\subsection{Pre-processing and Feature Extraction}
\label{sec:preliminary}
\subsubsection{Background Sensor Data}
\label{subsec:background}
The background sensors considered in this study are accelerometer, gravity sensor, gyroscope, linear accelerometer, and magnetometer. 
As different devices are equipped with sensors with a different acquisition rates (generally distributed over three values: 50Hz, 100Hz, and 200Hz), in order to equalize the acquisition rates across subjects in the HuMIdb database, a variable down-sampling ratio $D$ is used, based on $f_{s}$, the frequency computed for each acquisition session based on the timestamp information associated to each sample ($D = 1$ for $f_{s} < 75Hz$, $D = 2$ for $75Hz \leq f_{s} < 150Hz$, $D = 4$ otherwise). We adopt this strategy to remove potential sources of bias related to the device rather than to the biometric information across different subject data. Moreover, we obtain a final frequency value in line with the literature, accounting for the typical trade-off between retaining a sufficient amount of information and the constraints of mobile hardware resources (energy and computation) \cite{Abuhamad}.
In this operation, we take the first value every $D$ samples. To minimize the impact of noise, data normalization is performed by subtracting the mean and dividing by the standard deviation of each time domain signals per axis per session. Following \cite{Deb2019}, the Fast Fourier Transform is extracted from the raw \textit{x}, \textit{y}, \textit{z} data. Additionally, we also include the first- and second-order derivatives. Their computation is in fact lightweight and their inclusion in the system proved to be beneficial.
Features pertaining to a single timestamp are arranged into a 12-dimensional vector (4-dimensional in the case of the gravity sensor, as in this case the sensor output is one-dimensional): 
\begin{center}
[\textit{x}, \textit{y}, \textit{z}, \textit{x'}, \textit{y'}, \textit{z'}, \textit{x''}, \textit{y''}, \textit{z''}, \textit{fft(x)}, \textit{fft(y)}, \textit{fft(z)}]
\end{center}

\subsubsection{Touch Data} 
\label{subsec:touch}
The tasks considered within this study are keystroke, scroll up, scroll down, drawing the number ``8" with the finger, and tapping. The keystroke data were acquired in a fixed-text scenario corresponding to the sentence ``En un lugar de la Mancha, de cuyo nombre no quiero acordarme", in Spanish for all users, being it the primary language of around 85\% of them. All mistakes made by the subject while typing were acquired. Finally, for the tapping task, the subjects had to tap on a series of buttons in predetermined locations of the screen. In the case of keystroke, the raw data consist in the timestamp and the value of the key pressed. Therefore, the features considered are the inter-press time and the normalized value of the ASCII code, as in such fixed-text scenario no user sensitive information is contained in the raw text. For the remaining touch tasks, the raw data consist in the spatial \textit{x} and \textit{y} coordinates of the screen, and the pressure \textit{p} applied. The preprocessing pipeline is similar to the case of background sensor signals, including first- and second-order derivatives and the FFT, but without down-sampling. The \textit{x} and \textit{y} data are divided by the height and width values of the screen of the particular device to avoid potential sources of bias across devices, and the pressure \textit{p} is normalized similarly to background sensor data. 

\subsection{Learning Architecture}
\label{sec:learning_architecture}
The core of the authentication system is an LSTM RNN, a DL architecture able to capture long-term dependencies in time domain sequences \cite{8259229}. The network developed consists in two 64-unit layers based on a \textit{tanh} activation function. Batch normalization, and a dropout rate of 0.5 between each layer and a recurrent dropout rate of 0.2 in each layer are implemented to avoid overfitting. Such hyper-parameter configuration was chosen after preliminary tests.
\par Consecutive data samples were arranged in $M$-sample time windows (with zero-padding if needed). Such time windows are at the foundation of the proposed system as each of them represents the biometric information unit provided to the network, for training and testing purposes. The value of $M$ depends on the modality. In the case of background sensors, $M = 150$. Consequently, given the down-sampling performed at a variable rate bringing all background sensor signals to around 50Hz (see Sec. \ref{subsec:background}), each time window lasts 3 seconds. For the scrolling tasks, the sequence length is set to $M = 100$, whereas for tapping $M = 15$, as such gestures are generally shorter and the number of samples acquired is lower.
\par For a given \textit{M}-sample time window, the model outputs an embedding, i.e. an \textit{E}-dimensional array of real values that provides a compact representation of the salient features lying within the data ($E = 64$). The system is developed with the objective of mapping time windows of the same subject to produce representation close to each other in the embedding space, whereas embeddings belonging to different subjects should be distant to each other.

\subsection{Training Approach}
\par A different unimodal network is trained for each of the ten modalities (5 touch tasks and 5 background sensors). For each different background sensor data, models are trained by extracting time windows evenly from the different touch tasks. During the initial stage of the development of the system, this strategy led in fact to the best performance in terms of EER, compared to having an independent model for every \textit{task-modality} combination. In such way, the model is able to learn more general and robust features.

\subsection{Triplet Loss Function} 
\label{subsec:triplet_loss}
In the context of neural networks, triplet loss is an extension of contrastive loss that allows learning from positive and negative comparisons at the same time, whereas in constrastive loss, only one comparison at a time is possible \cite{JMLR:v10:weinberger09a}. A triplet consists in an ordered sequence of three separate time windows belonging to two different classes: the Anchor (A) and the Positive (P) are different time windows from the same subject, whereas Negative (N) is a time window from a different subject. The triplet loss function is defined below: 
\begin{equation*}
\mathcal{L}_{TL} = max\{0, d^2(\mathbf{v}_{A}, \mathbf{v}_{P}) - d^2(\mathbf{v}_{A}, \mathbf{v}_{N}) + \alpha \}
\end{equation*}
where $\alpha$ is the margin between positive and negative pairs and $d$ is the Euclidean distance between \textit{anchor}-\textit{positive} $(\mathbf{v}_{A}$-$\mathbf{v}_{P})$ pairs and \textit{anchor}-\textit{negative} $(\mathbf{v}_{A}$-$\mathbf{v}_{N})$ pairs. Triplet loss aims to minimize the distance between embedding vectors from the same class $(d^2(\mathbf{v}_{A}, \mathbf{v}_{P}))$, and to maximize it for different class embeddings $(d^2(\mathbf{v}_{A}, \mathbf{v}_{N}))$ within a single operation.

\subsection{Fusion of Modalities}
\label{subsec:fusion_of_modalities}
Once the training process of the different individual biometrics system is performed, we carry out a multimodal fusion between them. Among the several possible data fusion strategies, the fusion between modalities is performed at score level \cite{ROSS20032115}. In other words, the fusion is a combination of the scores, in terms of Euclidean distances between embeddings obtained from the different modalities available within the same time window. Fusion at data-level is a different approach adopted in the literature for authentication systems based on behavioral biometrics \cite{Abuhamad}. In such case, the different modalities are fused at an earlier stage, before being fed to the classifier. While this approach allows to train only one model which could learn more directly the possible relations between the different modalities, our score level fusion provides the system with more modularity, as different models can be integrated at different times, as long as their output are in the same form. In this way, individual models can be optimized individually, creating margin for improvement.
\par The scores produced by the models proposed in the current study lie in a close range, and the approach for the current experiments is a weighted summation of the scores, according to the formula:
\begin{equation*}
{S}_{W} = \sum_{n=1}^{N} \hat{w}_{n}*s_{n} 
\end{equation*}
where, for every specific subset of modalities, $n$ is the modality index, $N$ is the number of modalities, $\hat{w}_{n}$ is the modality normalized weight based on its performance on the validation set, and $s_{n}$ is the initial modality score. 
\par For each of the touch tasks, six modalities are combined: touch data from each task (keystroke, scroll up, scroll down, drawing an 8 with the finger, tapping), and the five background sensors (accelerometer, gravity sensor, gyroscope, linear accelerometer, magnetometer), leading to 63 different fusion combinations. 

\begin{table}[t]
\vspace{-5mm}
\caption{The mean number and the standard deviation of time windows obtained from the 3 enrolment sessions averaged over the different modalities across the different tasks.}

\centering
\small
\label{tab:netw}
\begin{tabular}{l l l}
\multicolumn{1}{l}{\textbf{Task}} & \multicolumn{1}{l}{\textbf{Mean}} & \multicolumn{1}{l}{\textbf{Std Dev}} \tabularnewline
\hline
\multicolumn{1}{l}{\textbf{Keystroke}} & \multicolumn{1}{c}{27.54} &
\multicolumn{1}{c}{18.00} \tabularnewline
\multicolumn{1}{l}{\textbf{Scroll Up}} & \multicolumn{1}{c}{3.54} & 
\multicolumn{1}{c}{6.29} \tabularnewline
\multicolumn{1}{l}{\textbf{Scroll Down}} & \multicolumn{1}{c}{1.98} &
\multicolumn{1}{c}{2.09} \tabularnewline
\multicolumn{1}{l}{\textbf{Drawing ``8"}} & \multicolumn{1}{c}{1.01} & \multicolumn{1}{c}{0.09} \tabularnewline
\multicolumn{1}{l}{\textbf{Tap}} & \multicolumn{1}{c}{1.27} & \multicolumn{1}{c}{1.60} \tabularnewline
\hline
\end{tabular}
\vspace{-5mm}
\end{table}

\begin{table*}[h!]\caption{Results in terms of EER (\%) of the different individual modalities during each task.}
\centering
\small
\label{tab:unimodal_moda}
\begin{tabular}{l l|l|l|l|l|l|}
\multicolumn{7}{ c }{\textbf{Individual Modalities}}\tabularnewline
\multicolumn{1}{ l }{Task} & \multicolumn{1}{ c }{Specific Task} & \multicolumn{1}{l}{Accelerometer} & \multicolumn{1}{l}{Gravity Sensor} & \multicolumn{1}{l}{Gyroscope} & \multicolumn{1}{l}{Lin. Accelerometer} & \multicolumn{1}{l}{Magnetometer} \tabularnewline
\Xhline{2\arrayrulewidth}
\textbf{Keystroke} & \multicolumn{1}{c}{\textbf{12.19}} & \multicolumn{1}{c}{19.23} & \multicolumn{1}{c}{29.18} & \multicolumn{1}{c}{26.27} & \multicolumn{1}{c}{17.05} & \multicolumn{1}{c}{14.03} \tabularnewline
\hline
\textbf{Scroll up} & \multicolumn{1}{c}{23.93} & \multicolumn{1}{c}{23.79} & \multicolumn{1}{c}{28.28} & \multicolumn{1}{c}{32.24} & \multicolumn{1}{c}{22.31} & \multicolumn{1}{c}{\textbf{15.23}} \tabularnewline
\hline
\textbf{Scroll down} & \multicolumn{1}{c}{23.08}& \multicolumn{1}{c}{19.22} & \multicolumn{1}{c}{23.08} & \multicolumn{1}{c}{26.15} & \multicolumn{1}{c}{20.11} & \multicolumn{1}{c}{\textbf{14.56}} \tabularnewline
\hline
\textbf{Tap} & \multicolumn{1}{c}{27.77}& \multicolumn{1}{c}{19.12} & \multicolumn{1}{c}{26.15} & \multicolumn{1}{c}{26.94} & \multicolumn{1}{c}{19.22} & \multicolumn{1}{c}{\textbf{12.46}} \tabularnewline
\hline
\textbf{Drawing ``8"} & \multicolumn{1}{c}{\textbf{14.52}}& \multicolumn{1}{c}{22.41} & \multicolumn{1}{c}{33.96} & \multicolumn{1}{c}{33.07} & \multicolumn{1}{c}{22.97} & \multicolumn{1}{c}{18.59} \tabularnewline
\hline
\multicolumn{2}{l}{\textbf{Average of Background Sensors}}& \multicolumn{1}{c}{\textbf{20.76}} & \multicolumn{1}{c}{\textbf{28.13}} & \multicolumn{1}{c}{\textbf{28.93}} & \multicolumn{1}{c}{\textbf{20.33}} & \multicolumn{1}{c}{\textbf{14.98}} \tabularnewline
\end{tabular}
\vspace{-2mm}
\end{table*}

\begin{table*}[t]\caption{Results in terms of EER (\%) of the three best subsets originated from the fusion (non-weighted) of the different individual modalities for each task.}
\centering
\small
\label{tab:multimodal_moda}
\begin{tabular}{l|l|l|l|l|l|l|l|l|l|}
\multicolumn{7}{ c }{\textbf{Fusion of Modalities}}\tabularnewline
\multicolumn{1}{l}{Task} & 
\multicolumn{1}{l}{Subset \#1} & \multicolumn{1}{l}{EER (\%)} &
\multicolumn{1}{l}{Subset \#2} & \multicolumn{1}{l}{EER (\%)} &
\multicolumn{1}{l}{Subset \#3} & \multicolumn{1}{l}{EER (\%)} 
\tabularnewline
\Xhline{2\arrayrulewidth}
\multicolumn{1}{l}{\textbf{Keystroke}} & 
\multicolumn{1}{l}{K, L, M} & \multicolumn{1}{c}{\textbf{4.62}} &
\multicolumn{1}{l}{K, A, M} & \multicolumn{1}{c}{5.21} &
\multicolumn{1}{l}{K, A, L, M} & \multicolumn{1}{c}{5.38} 
\tabularnewline
\hline
\multicolumn{1}{l}{\textbf{Scroll up}} & 
\multicolumn{1}{l}{SU, A, Gr, L, M} & 
\multicolumn{1}{c}{\textbf{8.55}} &
\multicolumn{1}{l}{SU, A, Gr, M} & \multicolumn{1}{c}{8.63} &
\multicolumn{1}{l}{SU, A, M} & \multicolumn{1}{c}{9.11} 
\tabularnewline
\hline
\multicolumn{1}{l}{\textbf{Scroll down}} & 
\multicolumn{1}{l}{SD, A, Gr, L, M} & \multicolumn{1}{c}{\textbf{6.87}} &
\multicolumn{1}{l}{SD, A, Gy, L, M} & \multicolumn{1}{c}{6.92} &
\multicolumn{1}{l}{SD, A, L, M} & \multicolumn{1}{c}{7.55} 
\tabularnewline
\hline
\multicolumn{1}{l}{\textbf{Tap}} &
\multicolumn{1}{l}{A, M} & \multicolumn{1}{c}{\textbf{7.69}} &
\multicolumn{1}{l}{A, Gr, L, M} & \multicolumn{1}{c}{7.80} &
\multicolumn{1}{l}{A, Gy, L, M} & \multicolumn{1}{c}{8.30} 
\tabularnewline
\hline
\multicolumn{1}{l}{\textbf{Drawing ``8"}} & 
\multicolumn{1}{l}{TD, A, Gr, L, M} & \multicolumn{1}{c}{\textbf{7.85}} &
\multicolumn{1}{l}{TD, A, Gr, M} & \multicolumn{1}{c}{8.46} &
\multicolumn{1}{l}{TD, A, M} & \multicolumn{1}{c}{8.51}
\tabularnewline
\hline
\multicolumn{7}{p{\dimexpr\linewidth-10\tabcolsep\relax}}{\footnotesize Acronyms of Tasks: K = Keystroke, SD = Scroll Down, SU = Scroll Up, T = Tap, TD = Touch Draw. Acronyms of Background Sensors: A = Accelerometer, Gr = Gravity Sensor, Gy = Gyroscope, L = Linear Accelerometer, M = Magnetometer.}
\end{tabular}
\vspace{-2mm} 
\end{table*}

\begin{table}[h!]\caption{Results in terms of EER (\%) of the three best subsets originated from the weighted fusion of the different individual modalities for each task.}
\centering
\small
\label{tab:multimodal_moda_fusi}
\begin{tabular}{l|l|l|}
\multicolumn{3}{ c }{\textbf{Fusion of Modalities}}\tabularnewline
\multicolumn{1}{l}{Task} & 
\multicolumn{1}{l}{Subset \#1} & \multicolumn{1}{l}{EER (\%)} 
\tabularnewline
\Xhline{2\arrayrulewidth}
\multicolumn{1}{l}{\textbf{Keystroke}} & 
\multicolumn{1}{l}{K, A, M} & \multicolumn{1}{c}{\textbf{3.96}}
\tabularnewline
\hline
\multicolumn{1}{l}{\textbf{Scroll up}} & 
\multicolumn{1}{l}{SU, A, Gr, L, M} & \multicolumn{1}{c}{\textbf{8.42}}
\tabularnewline
\hline
\multicolumn{1}{l}{\textbf{Scroll down}} & 
\multicolumn{1}{l}{SD, A, Gr, Gy, L, M} & \multicolumn{1}{c}{\textbf{6.94}} 
\tabularnewline
\hline
\multicolumn{1}{l}{\textbf{Tap}} &
\multicolumn{1}{l}{T, A, M} & \multicolumn{1}{c}{\textbf{7.69}} 
\tabularnewline
\hline
\multicolumn{1}{l}{\textbf{Drawing ``8"}} & 
\multicolumn{1}{l}{TD, A, M} & \multicolumn{1}{c}{\textbf{7.69}}
\tabularnewline
\hline
\multicolumn{3}{p{\dimexpr\linewidth-7\tabcolsep\relax}}{\footnotesize Acronyms of Tasks: K = Keystroke, SD = Scroll Down, SU = Scroll Up, T = Tap, TD = Touch Draw. Acronyms of Background Sensors: A = Accelerometer, Gr = Gravity Sensor, Gy = Gyroscope, L = Linear Accelerometer, M = Magnetometer.}
\end{tabular}
\vspace{-6mm}

\end{table}
\vspace{-3mm}
\section{HuMIdb Database}
\vspace{-3mm}
\label{sec:dataset}
The Human Machine Interaction database (HuMIdb) is a freely available database including data acquired by 14 mobile sensors during natural human-mobile interaction of 600 subjects with a total of 179 different device models \cite{Acien2020b}. The acquisition data was completed across five sessions separated by a 24-hour gap at least, in order to account for intra-subject variability. The participants were asked to install an Android application on their own smartphone and to complete eight tasks in an unsupervised scenario. Regarding the age distribution, 25.6\% of the users were younger than 20 years old, 49.4\% are between 20 and 30 years old, 19.2\% between 30 and 50 years old, and the remaining 5.8\% are older than 50 years old. Regarding the gender, 66.5\% of the participants were males, 32.8\% females, and 0.7\% others. The subjects were recruited from 14 countries (52.2\% European, 47.0\% American, 0.8\% Asian). The tasks are fixed-text keystroke, scroll up and down gestures, tap gestures, draw a circle and a cross in the air with the smartphone, pronouncing a fixed sentence, and draw 0-9 digits with the finger on the touchscreen, and a swipe right button. The background sensors included are accelerometer, linear accelerometer, gyroscope, magnetometer, orientation, proximity, gravity, light, GPS, Wi-Fi, Bluetooth, and the microphone.

\vspace{-3mm}
\section{Experimental Protocol}
\label{sec:experimental_protocol}
Data subjects lacking one out of the five acquisition sessions of HuMIdb or presenting acquisition errors (i.e., arrays of zeros) are discarded. 
Training, validation, and test sets are formed from the remaining users without any overlap. For each of the modalities, the training set has a variable number of users (on average 246), whereas the validation and the test sets include the exact same subjects for all the different modalities (65 users in each of the two sets). Concerning the training hyper-parameters, the batch size is 512, the learning rate is set to 0.05, the Adam optimizer is used with $\beta_{1} = 0.9$, $\beta_{2} = 0.999$ and $\epsilon = 10^{-8}$. The models are built in \texttt{Keras-Tensorflow}.
\par During the training and validation of the networks, the time windows are extracted considering a random initial time instant across the entire sequence timeline of each session, always allowing for full $M$-sample time windows. To evaluate the performance of the networks, given each subject, their time windows are compared to all other subjects'. In other words, given $N$ subjects, for each subject there will be $N-1$ impostors. The time windows from the first three acquisition sessions (out of five) are considered as enrollment data, whereas the time windows from the remaining last 2 sessions are used for verification. For background sensors, all the available time windows per enrolment session are extracted considering an overlap time window of 50 samples, corresponding to roughly 1 second, whereas the duration of the time windows $M$ is still 150 samples. For touch tasks, we consider only one time window, given the brevity of such time sequences. Table \ref{tab:netw} shows the number of time windows obtained from the 3 enrolment sessions averaged over the different modalities across the different tasks. Depending on the duration of the specific task, the number of time windows obtained can change significantly. 
\par At verification time, one time window per each of the 2 remaining separate acquisition sessions is obtained. Then, two score values (one per verification session) consist in the distance value averaged over all the available enrolment time windows from the 3 enrolment sessions. Finally, for every subject 2 genuine distance values is calculated (one for each verification time window) and $2 \times (N-1)$ impostor scores. For these operations, the background sensor data considered during the evaluation phase is restricted within the time range given by the first and last sample of their respective simultaneous touchscreen task data.

\vspace{-3mm}
\section{Experimental Results}
\label{sec:experimental_results}
\subsection{Individual Modalities}
\label{subsec:indi}
Table \ref{tab:unimodal_moda} shows the results obtained on the test set considering each modality individually. Each row presents the results pertaining to a single task, and the corresponding background sensor data acquired simultaneously in the following columns. The metric chosen to evaluate the system performance is the EER. The touch task results are distributed in three groups. The best performing group includes keystroke and drawing the number ``8" with the finger, achieving respectively 12.19\% and 14.52\%. Then, scroll gestures produce a score around 23\% with little difference given by their direction. Finally, an EER score of 27.77\% is obtained during the tap task. Such distributions show the ability of the systems to extract more discriminative information from typing and drawing compared to swiping or tapping. The last row of Table \ref{tab:unimodal_moda} shows the EER scores achieved by each background sensor averaged over the tasks. The magnetometer consistently proves to be the best performing sensor (14.98\%), followed by lin. accelerometer (20.33\%), accelerometer (20.76\%), gravity sensor (28.13\%) and gyroscope (28.93\%). Moreover, there is no significant correlation between the scores obtained by background sensors and the duration of the corresponding task in terms of number of time windows extracted (see Table \ref{tab:netw}).

\begin{figure*}[!htb]
\centering
\phantomsection\label{fig:detcurves}
\minipage{0.34\textwidth}
  \includegraphics[width=0.97\linewidth]{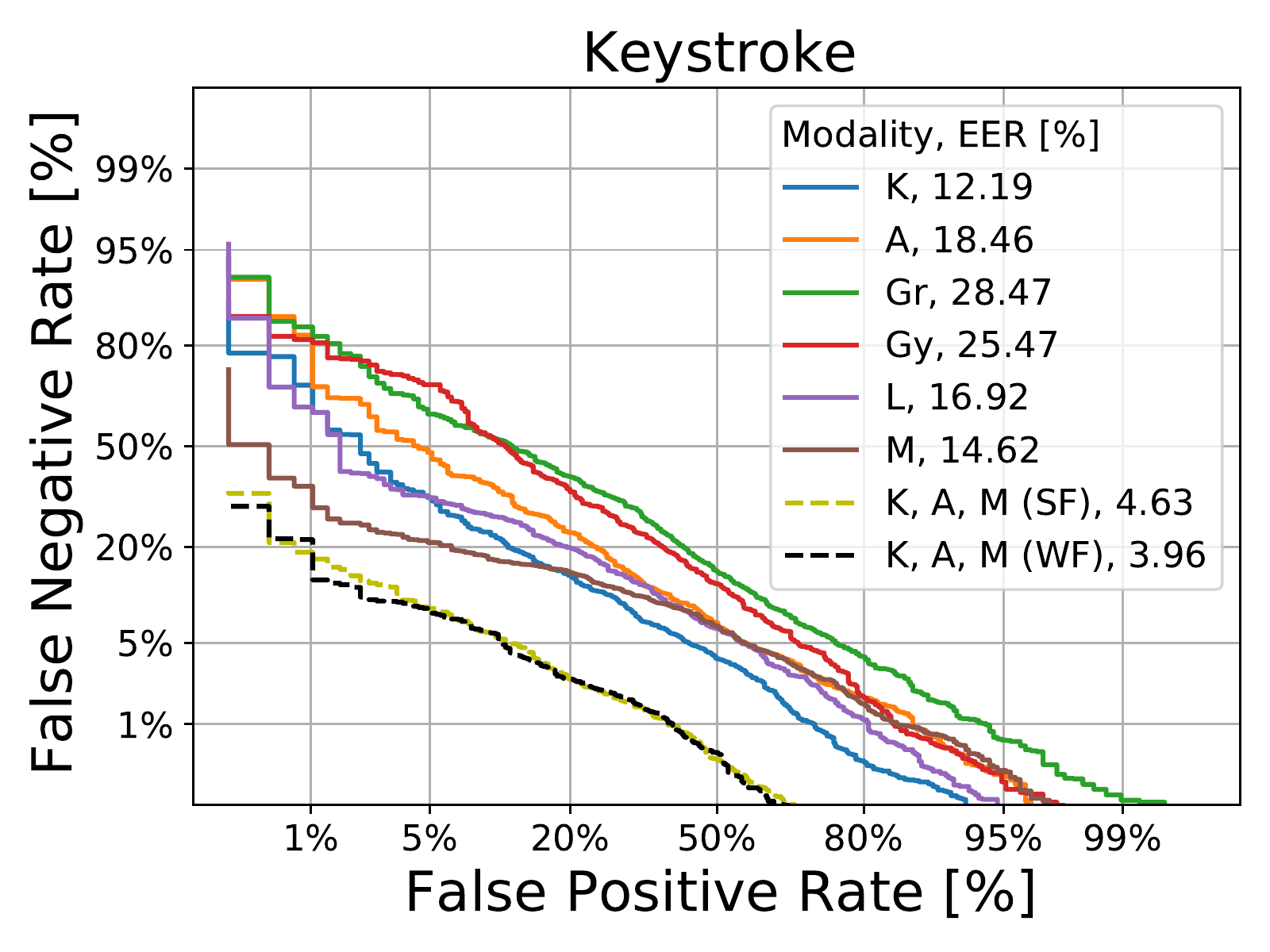}
\endminipage
\minipage{0.34\textwidth}
  \includegraphics[width=0.97\linewidth]{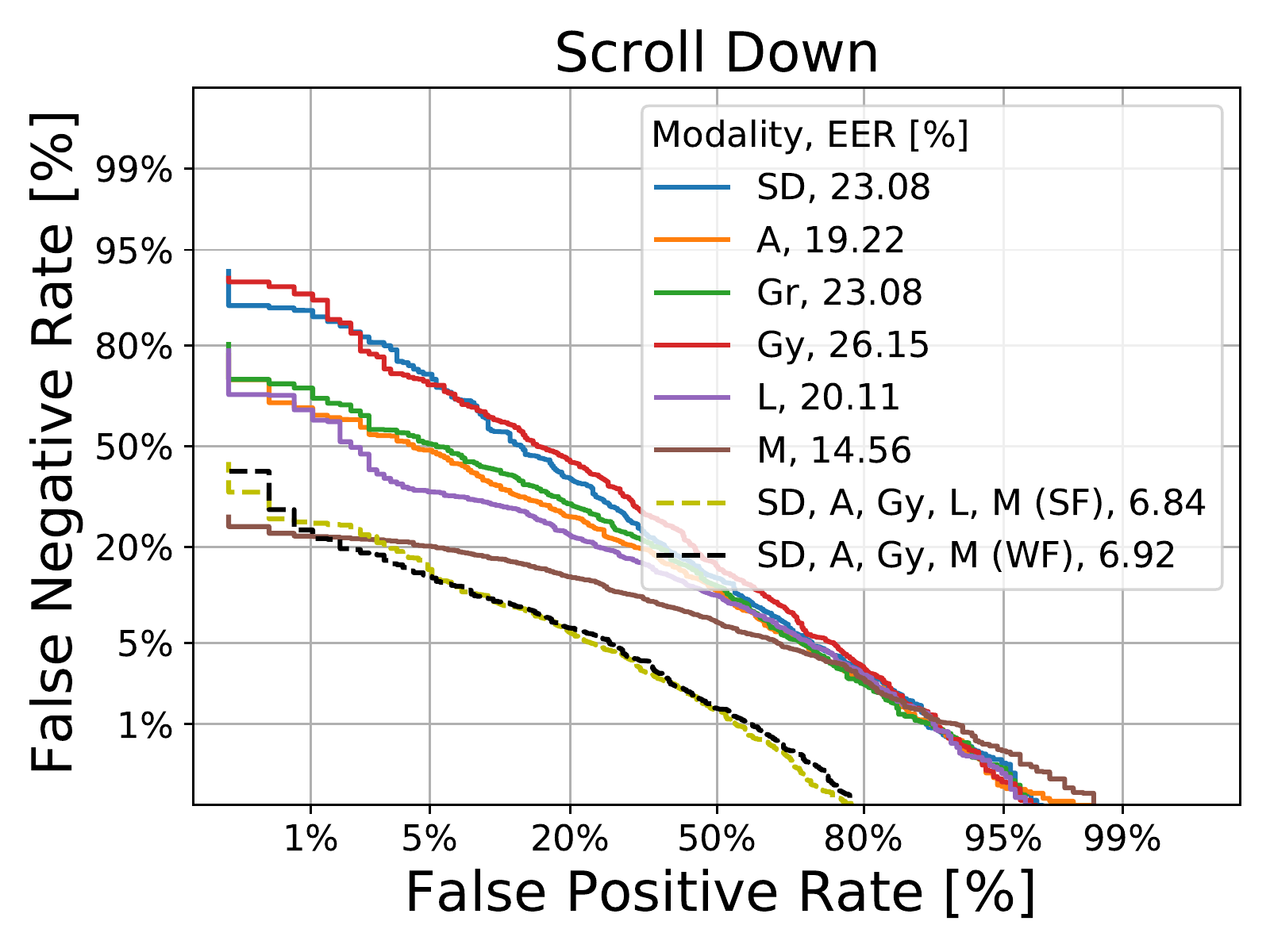}
\endminipage
\minipage{0.34\textwidth}
  \includegraphics[width=0.97\linewidth]{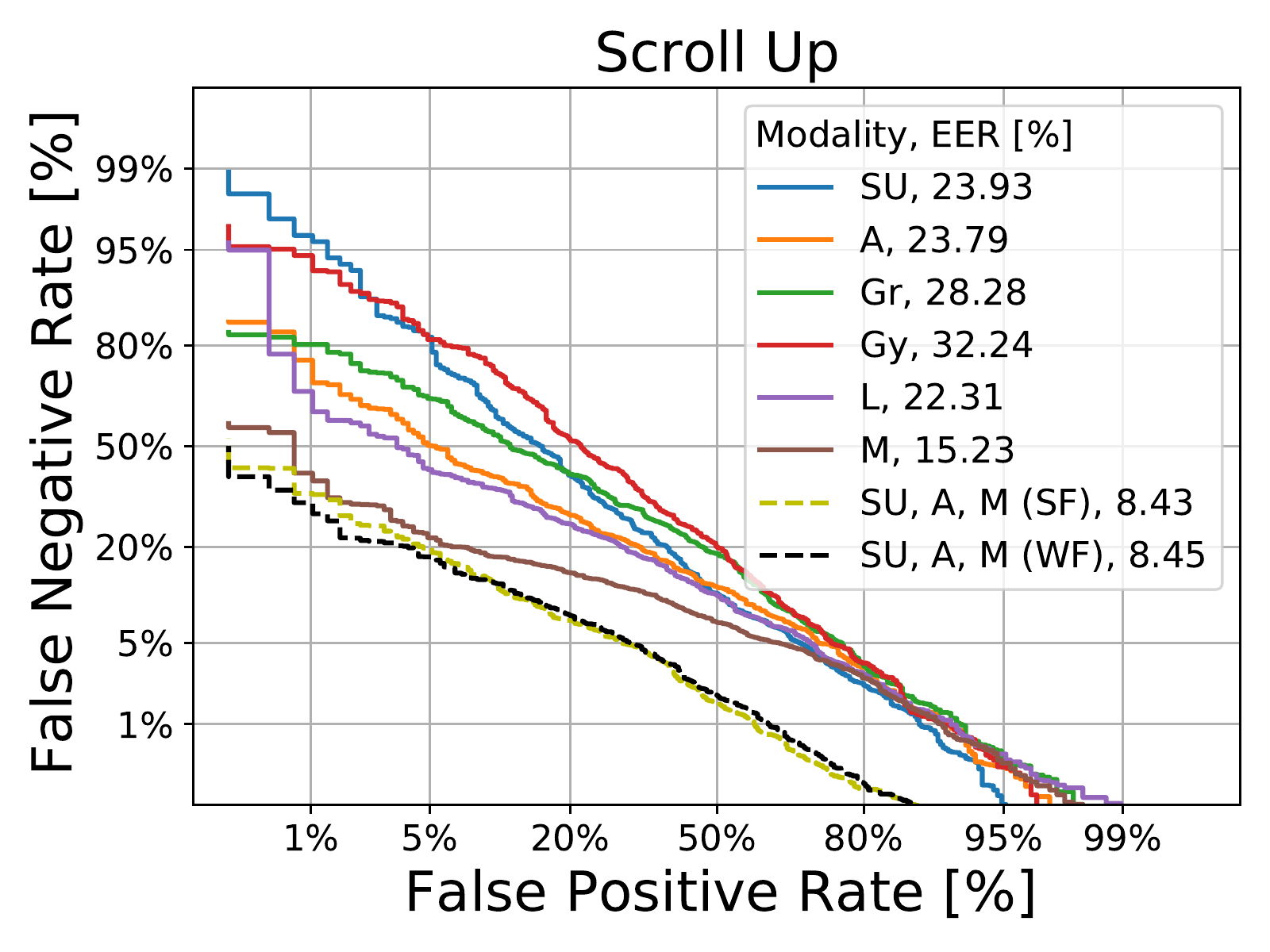}
\endminipage
\vspace{-3mm}
\end{figure*}
\begin{figure*}[!htb]
\centering
\minipage{0.34\textwidth}
  \includegraphics[width=0.97\linewidth]{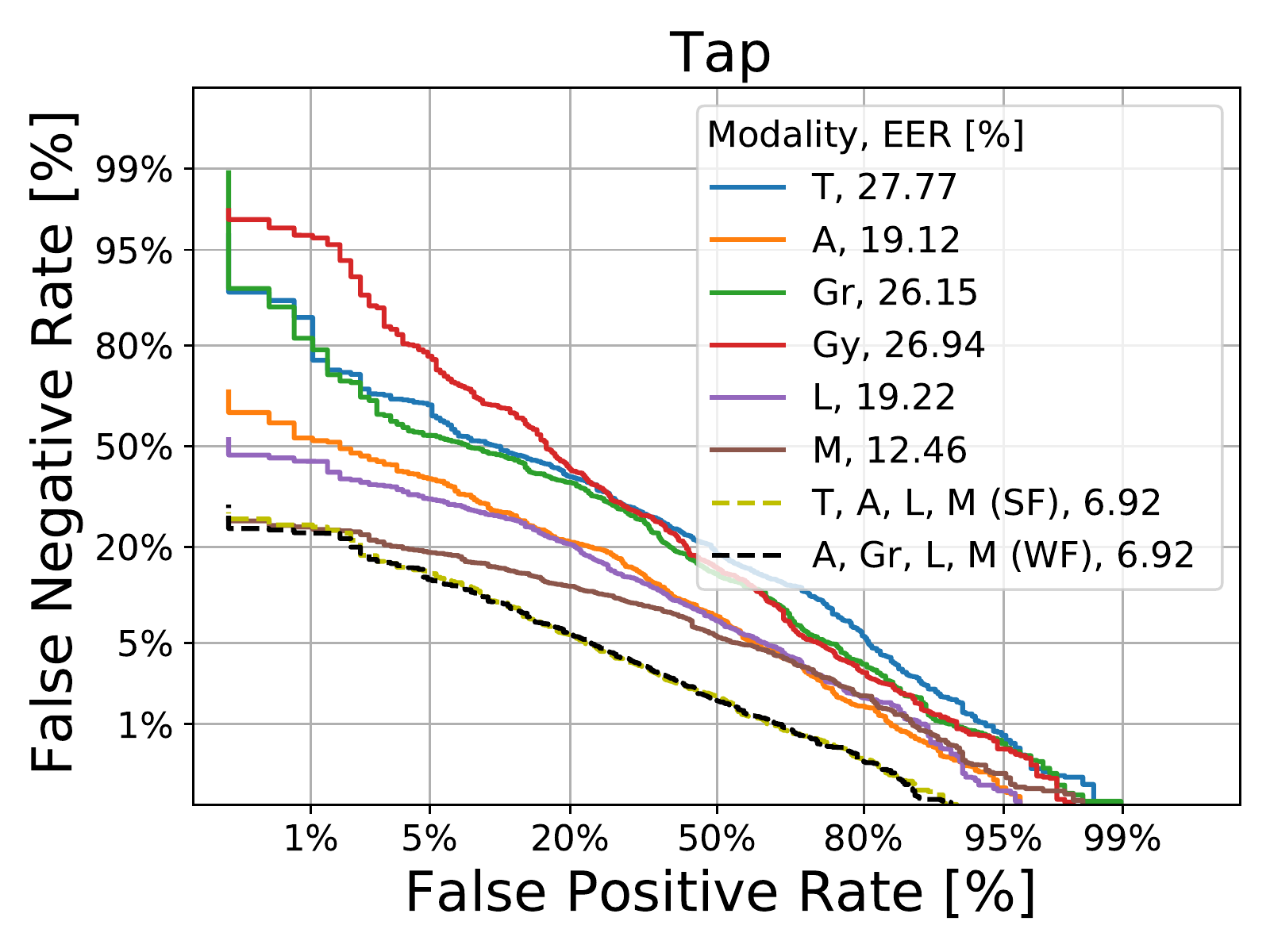}
\endminipage
\minipage{0.34\textwidth}
  \includegraphics[width=0.97\linewidth]{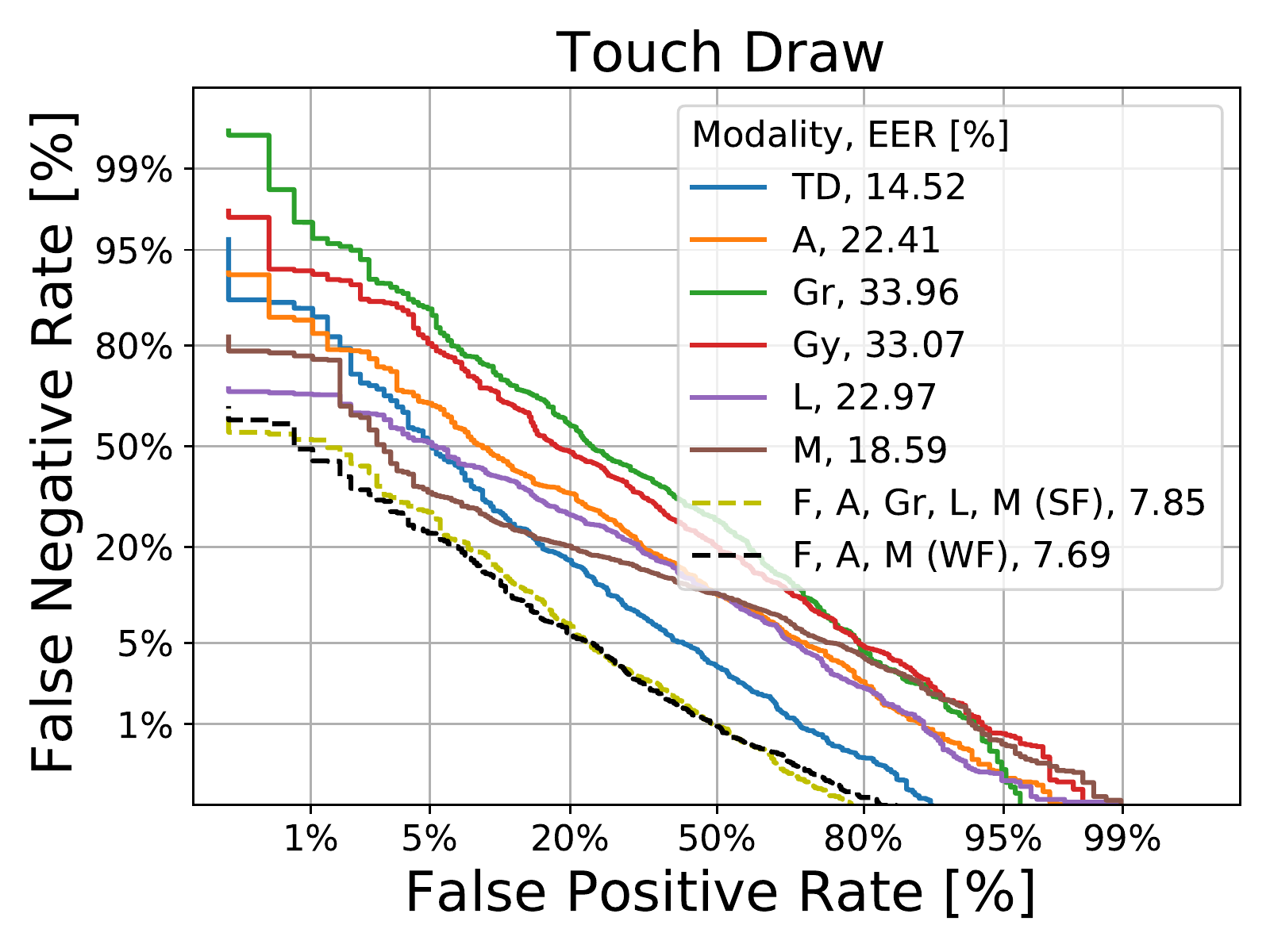}
\endminipage
\caption{The DET curves calculated for the keystroke task are displayed above. Each of the graph refers to a task and it contains the DET curve of each modality individually and of the best fusion combination subset for the case of simple fusion (SF) and weighted fusion (WF). To have the same amount of enrolment data, the displayed results are obtained considering one time window per enrolment session.}
\vspace{-5mm}
\end{figure*}

\subsection{Fusion of Modalities}
\label{subsec:fusion}

\par In Table \ref{tab:multimodal_moda}, the best three subsets originated from the fusion of modalities are included. In any case, the improvement in the performance of the system due to the fusion of modalities is significant. The individual modalities achieve an average EER of 21.46\%, whereras the average EER of the best modality combinations is 7.12\%, i.e., a relative error reduction of 66.82\% EER. 
The best performance is achieved for the task of keystroke with a fusion of the touch information and the accelerometer, gravity sensor, and magnetometer (K, L, M), reaching 4.62\% EER. In this case, the EER produced by the fusion of modalities is nearly one third of the one achieved with the touch data only. The second best performance is given by the gesture of scroll down (SD, A, Gr, L, M), i.e., 6.87\% of EER. 
During the scroll up task the best score obtained (SU, A, Gr, L, M) is around 1.7 percentage points higher than its counterpart, whereas drawing of the number ``8" with the finger (TD, A, Gr, L and M) and tapping (A, M) lead to similar results, slightly less than 8\% EER. With regard to the latter, it is interesting to point out how the three best combinations of modalities never contain the touch data. Such trend occurs only for this task, which, however, is the worst performing individual touch task. In all the cases of the table, however, the fusion of the modalities provides EER values below 10.00\%. 
Table \ref{tab:multimodal_moda_fusi} shows the results obtained with weighted coefficients computed on the validation set. Generally, they yield lower error rates, reaching less than 4\% for the best modality subset during the keystroke task. There are, however, few exceptions, which take place in the case of the task of scroll down and are due to the inevitable mismatch between the validation and the test set. Overall, the distribution of the scores is very similar compared to the non-weighted case. Figure \hyperref[fig:detcurves]{2} shows the Detection Error Trade-off (DET) curves for the five tasks considered. In each task the figure shows the DET curve for each individual model and for the two fusions considered. The benefits of the two fusion approaches are visible across all five tasks, even though the difference between the two is not very significant.

\subsection{Discussion}
\label{subsec:discussion}
\par Previous studies in the context of multimodal mobile passive authentication based on background sensors achieve better authentication results (see Sec. \ref{sec:related_work}). In such cases \cite{Deb2019, Abuhamad}, however, the time windows are extracted as long as the dedicated data acquisition app is running in the background, with loose requirements in terms of sensor activity. A huge amount of information per user is acquired, including many instances without any usable information. In the context of mobile authentication systems based on behavioral biometrics, it is crucial to decorrelate the notions of \textit{user} and \textit{device} \cite{Neverova2016, das2015exploring}. The amount of biometric information contained on average in a time window randomly sampled from the entire time the device is on is presumably less than in dense gesture-centered dedicated sessions, designed to mimic the most salient traits of mobile Human-Computer Interaction (HCI). Consequently, it would be interesting to investigate how much of the authentication effectiveness should be attributed to the models extracting and recognizing features belonging to the device rather than the user. Additionally, the sampling frequency of 1 Hz used in \cite{Deb2019} seems too low to acquire sufficient discriminative behavioral biometric information in 3-second time windows. Therefore, in this scenario, it would be difficult to reach a global and significant conclusion from the comparison of such systems, given the different approaches, scopes and the usage of self-collected non-public databases. In any case, given its characteristics and completeness, we point out the lack of public databases similar to the HuMIdb. While aiming to tackle the challenge of multimodal mobile biometric passive authentication from a different angle, a down-side of the proposed approach is undoubtedly the scarcity of training data per user compared to related studies.

\vspace{-3mm}
\section{Conclusions}
\vspace{-1.5mm}
\label{sec:conclusions}

The current work has focused on an analysis of individual and multimodal behavioral biometric traits suitable for the application of mobile continuous authentication, with fusion at score level. 
The study carries out a first benchmark of HuMIdb, a novel publicly available database of over 5GB of wide ranging mobile data collected in an unsupervised scenario. The modalities considered are based on touchscreen and background sensor data. For every modality, a separate LSTM RNN with triplet loss was considered. 
Our results show that the best performing source is the fixed-text keystroke data for touchscreen data and the magnetometer for background sensors. However, the discriminative ability of the system is significantly enhanced by the fusion, typically reaching a 4\%-9\% EER range within approximately 3 seconds of interaction. 
A large and public database as HuMIdb could be very useful to compare systems proposed in the literature, possibly in the form of an open competition. 
\par Mobile authentication systems based on behavioral biometrics, while covering the aspects of security and usability described in Sec. \ref{sec:intro}, currently do not achieve the same authentication performance as their counterparts based on physiological biometrics, such as face or fingerprint. Nevertheless, this is a popular topic of research in the biometric community as a form of complementary technology or second factor in a 2-factor authentication (2FA).
In the real-world scenario of a theft, the impostor and the genuine user data originate from the same device, it would be interesting to assess the device fingerprinting due to sensor differences and calibration imperfections, and decorrelate user from device recognition. This could be investigated by collecting user data from the same acquisition mobile device.
\par Moreover, in the context of behavioral biometric authentication based on data acquired during dedicated acquisition sessions, a crucial aspect is given by the limited amount of data per user. To compensate the reduced capacity of acquiring a large database, future work includes the generation of synthetic data, which proved to be a powerful tool in related fields \cite{TOLOSANA2020131}.

\vspace{-3mm}
\section*{Acknowledgments}
\vspace{-3mm}
This project has received funding from the European Union’s Horizon 2020 research and innovation programme under the Marie Skłodowska-Curie grant agreement No 860315, and from Orange Labs. R. Tolosana and R. Vera-Rodriguez are also supported by INTER-ACTION (PID2021-126521OB-I00 MICINN/FEDER). 

\vspace{-3mm}
\bibliographystyle{elsarticle-num} 
\bibliography{main}

\end{document}